\documentclass[11pt,a4paper]{article}
\usepackage[hyperref]{ranlp2023}
\usepackage{times}
\usepackage{latexsym}

\usepackage{colortbl}

\usepackage{graphicx}
\usepackage{array,multirow}

\usepackage[acronym]{glossaries}

\usepackage{microtype}

\aclfinalcopy 

\title{On the Limitations of Large Language Models (\acrshort{llm}s): False Attribution}

\author{\\Tosin Adewumi*, Nudrat Habib, Lama Alkhaled, and Elisa Barney
\\
Machine Learning Group, EISLAB,
\\ Luleå University of Technology, Sweden \\
\large
{\fontfamily{pcr}\selectfont
firstname.lastname@ltu.se}
}

\newacronym{ml}{ML}{Machine Learning}
\newacronym{llm}{LLM}{Large Language Model}
\newacronym{sota}{SotA}{state-of-the-art}
\newacronym{iaa}{IAA}{inter-annotator agreement}
\newacronym{nlp}{NLP}{natural language processing}
\newacronym{ai}{AI}{artificial intelligence}
\newacronym{llama}{LLaMA}{Large Language Model Meta AI}
\newacronym{bert}{BERT}{Bidirectional Encoder Representations from Transformers}
\newacronym{svm}{SVM}{support vector machine}
\newacronym{gru}{GRU}{gated recurrent unit}
\newacronym{cnn}{CNN}{convolutional neural network}
\newacronym{t5}{T5}{Text-to-Text Transfer Transformer}
\newacronym{shi}{SHI}{Simple Hallucination Index}
\newacronym{gai}{GenAI}{generative artifical intelligence}
\newacronym{roberta}{RoBERTa}{Robustly optimized BERT pretraining Approach}
\newacronym{al}{AL}{Active Learning}
\newacronym{hf}{HF}{HuggingFace}
\newacronym{qa}{QA}{Question Answering}
\newacronym{meteor}{METEOR}{Metric for Evaluation of Translation with Explicit ORdering}
\newacronym{c4}{C4}{Colossal Cleaned Common Crawl}

\begin{document}
\maketitle
\begin{abstract}
In this work, we introduce a new hallucination metric - \acrfull{shi} and provide insight into one important limitation of the parametric knowledge of large language models (\acrshort{llm}s), i.e. false attribution.
The task of automatic author attribution for relatively small chunks of text is an important \acrshort{nlp} task but can be challenging.
We empirically evaluate the power of 3 open \acrshort{sota} \acrshort{llm}s in zero-shot setting (Gemma-7B, Mixtral 8x7B, and \acrshort{llama}-2-13B).
We acquired the top 10 most popular books of a month, according to Project Gutenberg, divided each one into equal chunks of 400 words, and prompted each \acrshort{llm} to predict the author.
We then randomly sampled 162 chunks per book for human evaluation, based on the error margin of 7\% and a confidence level of 95\%.
The average results show that Mixtral 8x7B has the highest prediction accuracy, the lowest \acrshort{shi}, and a Pearson's correlation (\textit{r}) of 0.724, 0.263, and -0.9996, respectively, followed by \acrshort{llama}-2-13B and Gemma-7B.
However, Mixtral 8x7B suffers from high hallucinations for 3 books, rising as high as a \acrshort{shi} of 0.87 (in the range 0-1, where 1 is the worst).
The strong negative correlation of accuracy and \acrshort{shi}, given by \textit{r}, demonstrates the fidelity of the new hallucination metric, which may generalize to other tasks.
We also show that prediction accuracies correlate positively with the frequencies of Wikipedia instances of the book titles instead of the downloads and we perform error analyses of predictions. 
We publicly release the annotated chunks of data and our codes to aid the reproducibility and evaluation of other models.\footnote{This paper was accepted for presentation by Recent Advances in NLP (RANLP) 2025 conference.}
\end{abstract}


\section{Introduction}
\label{intro}

\begin{table}[htbp]
\small
\caption{Output example of Mixtral 8x7B with input from a chunk of \textit{The Expedition of Humphry Clinker} by Tobias Smollet. (\footnotesize{Bold style for emphasis.})}\label{exampleout}
\centering
\begin{tabular}{| p{0.16\linewidth} | p{0.72\linewidth}| }
\hline
Prompt Input: & \cellcolor[HTML]{dbd9ce}Who is the author of this text: 'how--does the young gentleman stand for Haverford West?...he wheeled about; and going round the levee, spoke to every individual, with the most courteous familiarity;...As for the common business of the nation, it is carried on in a constant routine by the clerks of the different offices, otherwise the wheels of government'     \\
\hline
Mixtral 8x7B Output: & \cellcolor[HTML]{D3D3D3}The text you provided is an excerpt from a longer work, and it does not include any information about the author. The text appears to be a piece of fiction, describing a scene in which a duke is holding a levee and receiving visitors. \textbf{The author of this text is unknown in this context}.    \\
\hline
\end{tabular}
\end{table}

False attribution is the incorrect representation that someone or an entity is the author of a work when they are actually not \cite{carty1989copyright,alqurashi-etal-2025-bert}.
This problem is not only an ethical or moral issue but also a legal one in some cases, as it is considered illegal in some quarters \cite{hicks2014right,ikonomi2021has}.
This may also be observed in the automatic attribution of text by a \acrfull{llm} in what may be called (intrinsic) hallucination \cite{alkaissi2023artificial}.
Hallucination, in the context of \acrshort{ai}, is when a model confidently presents a piece of false information as a fact \cite{maynez-etal-2020-faithfulness,10.1145/3571730}.
This risk is a poignant issue in automatic annotation because of the increasing appeal to use automatic annotation (or labels) by \acrshort{llm}s due to the high cost of human annotation.
\acrshort{llm}s are large neural probabilistic models that are pretrained on large amounts of data (\textbf{including books}) through self-supervised learning to predict the next token and finetuned for downstream tasks \cite{radford2019language,brown2020language,adewumi2023procot}.
It appears many existing hallucination metrics are based on a binary format, such as factual or non-factual \cite{lee2022factuality,kang2024comparing}, yes or no,\footnote{docs.rungalileo.io/galileo/gen-ai-studio-products/guardrail-store/factuality} and other binary options \cite{li-etal-2023-halueval}.
This is inadequate and misleading, especially for a task such as \acrfull{qa}, as we believe a system should not be penalized for saying \textit{I don't know}, as in the example in Table \ref{exampleout}


In this work, our objective is to demonstrate, in \textbf{zero-shot setting}, the strengths and limitations of \acrshort{llm}s with regards to the task of author attribution for chunks of text and introduce a simple hallucination metric for their evaluation - \acrfull{shi}.
This work will provide valuable feedback to the research community for improving \acrshort{llm}s for more trustworthiness.
In order to answer our research question of "\textbf{how do recent open \acrshort{llm}s fare with regards to false attribution for short texts of books in zero-shot setting?}", we selected, as the dataset, the 10 most downloaded (or popular) books\footnote{for the month of March, 2024; at gutenberg.org/ebooks/bookshelf}, according to Project Gutenberg.
More details about the books are provided in Section \ref{method} and they include \textit{Pride and Prejudice}, \textit{Moby Dick}, \textit{Middlemarch}, \textit{The Adventures of Ferdinand Count Fathom}, \textit{The Expedition of Humphry Clinker}, \textit{The Adventures of Roderick Random}, \textit{History of Tom Jones}, \textit{A Doll’s House}, \textit{Crime and Punishment}, and \textit{Great Expectations}.

Three fairly recent, open-weight \acrfull{sota} \acrshort{llm}s (Gemma-7B, Mixtral 8x7B, and \acrshort{llama}-2-13B) were evaluated on equal chunks of the books before sampling from each book for human evaluation.
The results reveal that Mixtral 8x7B is the best model but it still suffers from high hallucination in some cases, with a \acrshort{shi} of 0.87 for one of the books.
Our contributions include the following:

\begin{itemize}
    \item We introduce a simple and novel hallucination metric for \acrshort{llm}s - \acrfull{shi} (pronounced \textit{shy}).
    This is important to build more trustworthy \acrshort{gai}.

    \item We publicly release the \acrshort{llm}-annotated chunks of data, which can be useful for author attribution tasks\footnote{https://github.com/LTU-Machine-Learning/llm\_limitation\_AA}

    \item We are the first, to the best of our knowledge, to demonstrate the false attribution problem in \acrshort{llm}s in a systematic way for chunks of books.
\end{itemize}

The rest of this paper is organized as follows.
In Section \ref{shi}, we explain the \acrshort{shi} metric.
Section \ref{method} discusses in detail the methods we employed in this work.
We present the results and analyses in Section \ref{res}.
Section \ref{related} discusses the related work from the literature.
We conclude with a summary and possible future work in Section \ref{conclude}.

\section{\acrfull{shi}}
\label{shi}
\acrshort{shi}, given by Equation \ref{eq:eq2}, differentiates unknown (\textit{u}) from incorrect (\textit{i}) predictions made by an \acrshort{llm}, unlike the typical binary (correct/incorrect) classes in author attribution tasks \cite{diederich2003authorship,savoy2016estimating} or hallucination metrics.
A binary hallucination metric takes the form of Equation \ref{eq:binary} and is too restrictive.
It forces an exaggeration of the evaluation, where the incorrect (\textit{i}\textsuperscript{*}) is a combination of the actual incorrect and the unknown cases.
The correct predictions are represented by \textit{c} in both equations.

\begin{equation}
\mathit{\acrshort{shi}} = 
\frac{i}{c + i + u}
\label{eq:eq2}
\end{equation}

\begin{equation}
\mathit{BinaryH} = \frac{i \textsuperscript{*}}{c + i\textsuperscript{*}} = 
\frac{i + u}{c + i + u}
\label{eq:binary}
\end{equation}


This important property of \acrshort{shi}, in considering the unknown (when the model is unable to give any prediction or when it explicitly says it's unsure), ensures it does not score the model positively.
This contrasts with the \textit{truthfulness} metric of TruthfulQA \cite{lin-etal-2022-truthfulqa} that assigns a score even when the model refuses to answer a question for any reason, the \textit{ensemble} of FactualityPrompt \cite{lee2022factuality} that is binary-based on factual and non-factual annotations, and HaluEval's accuracy \cite{li-etal-2023-halueval}, which is also binary-based on hallucinated or normal samples.
Furthermore, these metrics are tied to specific benchmarks or data about world facts, making them less flexible.
On the other hand, \acrshort{shi} can be applied to any task involving \acrshort{llm}s and is not dependent on any specific benchmark or dataset.

If we compare \acrshort{shi} to other standard metrics like Precision (\textit{P}) (Equation \ref{eq:precision}), Recall (\textit{R}) (Equation \ref{eq:recall}), \textit{F1} (Equation \ref{eq:f1}), \textit{Accuracy} (Equation \ref{eq:acc}) and the 
\textit{\acrfull{meteor}} (Equation \ref{eq:meteor}) \cite{banerjee2005meteor}, which may be used in hallucination evaluation \cite{chen2023hallucination,chang2024survey}, we may observe their limitation.
This is because such metrics are based on true positives (\textit{tp}), true negatives (\textit{tn}), false positives (\textit{fp}), and false negatives (\textit{fn}), none of which accounts for unknown cases.


\begin{equation}
\mathit{P} = 
\frac{tp}{tp + fp}
\label{eq:precision}
\end{equation}

\begin{equation}
\mathit{R} = 
\frac{tp}{tp + fn}
\label{eq:recall}
\end{equation}

\begin{equation}
\mathit{F1} = 
{2} . \frac{P . R}{P + R}
\label{eq:f1}
\end{equation}

\begin{equation}
\mathit{Accuracy} = 
\frac{tp + tn}{tp + tn + fp + fn}
\label{eq:acc}
\end{equation}

\begin{equation}
\mathit{\acrshort{meteor}}  = 
{10} . \frac{P . R}{R + 9.P} . {\left( 1 - penalty \right)}
\label{eq:meteor}
\end{equation}

\section{Methodology}
\label{method}

All the experiments were performed on an Nvidia DGX-1 node, with 8 x 40GB A100 GPUs, that runs on Ubuntu 22.04.
The 3 \acrshort{llm}s we evaluated are chat (or instruction-tuned) models of Gemma-7B-In, Mixtral 8x7B, and \acrfull{llama}-2-13B.
These models were used due to compute and time constraint.
They are sourced from the \acrfull{hf} hub \cite{wolf-etal-2020-transformers}.
We kept the default hyper-parameters in the \acrshort{hf} and set the maximum number of tokens for each to 1,200.
Table \ref{llms} provides a brief summary of the properties of the \acrshort{llm}s.
We follow previous work and use accuracy to report prediction performance \cite{luyckx2008authorship,mallen-etal-2023-trust}.

\begin{table}[h!]
\small
\centering
\caption{Properties of the \acrshort{llm}s}\label{llms}

\begin{tabular}{p{0.26\linewidth} | p{0.16\linewidth} | p{0.16\linewidth} | p{0.16\linewidth}}
\hline
\textbf{Properties} & \textbf{Gemma-7B} & \textbf{Mixtral 8x7B}  & \textbf{\acrshort{llama}-2-13B}\\
\hline
Parameters & 7B & 45B & 13B \\ \hline
Pretraining data & Web & Open Web & Public online data\\ \hline
Pretraining tokens & 6 T & unknown & 2 T \\ \hline
Context length & 8k & 32k  & 4k \\ \hline
Languages & English & Multi & Multi \\ \hline
\acrshort{hf} Leaderboard & 56.4 & 68.42 & 54.91 \\ \hline
License & Gemma & Apache 2.0 & \acrshort{llama}-2 \\ 
 \hline
\end{tabular}
\end{table}

\subsection{The Dataset}

The evaluation dataset is the 10 most downloaded (or popular) books, as mentioned in Section \ref{intro}.
They span different fiction genres.
The data statistics of the books are provided in Table \ref{tabpop}. 
We follow \citet{bevendorff2019generalizing} and \citet{hicke2023t5} and split each book into chunks of text of 400 words.
The last chunk for each book usually contains less than 400 words.

In order to further validate the popularity of the books, we verified the frequency of times the books are mentioned in 
Wikipedia\footnote{en.wikipedia.org/w/index.php?search=\&title=Special:Search\&profile= advanced\&fulltext=1\&ns0=1} (see Table \ref{tabpop}), in contrast to the approach of using monthly page views as a proxy as done by \citet{mallen-etal-2023-trust}.
This is because of the valid assumption that the more the instances in the training data of a model, the better the performance of the model \cite{brown2020language,zhou2017machine}.
We use Wikipedia since this is what was used in previous studies \cite{kandpal2023large,mallen-etal-2023-trust} and it is part of the pretraining data for the 3 \acrshort{llm}s in Table \ref{llms} and others \cite{raffel2020exploring,brown2020language}.
The Wikipedia frequencies will be useful in the analysis of the results of this work.


\begin{figure*}[htbp]
\centering
\includegraphics[width=0.9\textwidth]{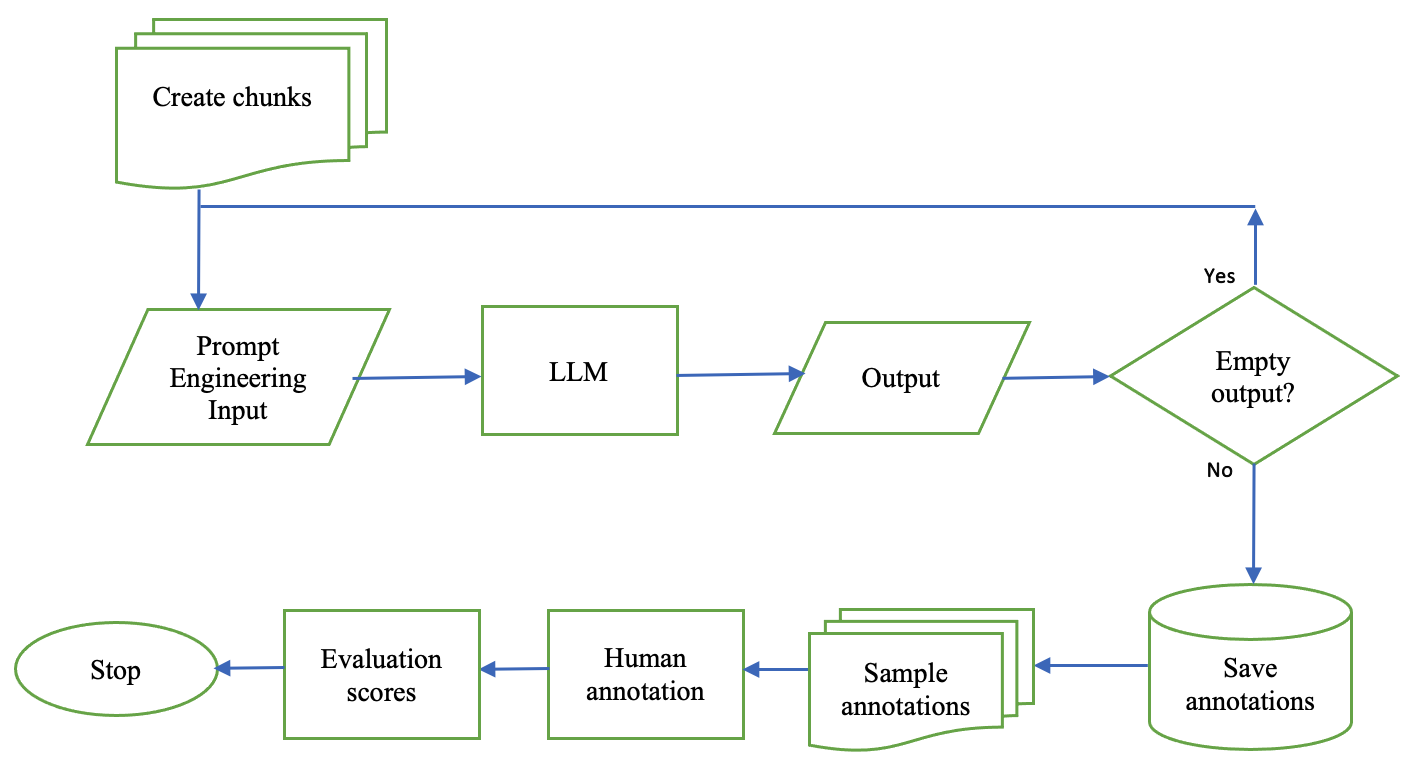}
\caption{Prediction lifecycle}
\label{annot_method}
\end{figure*}

\begin{table}[h!]
\tiny
\centering
\caption{The 10 most popular books for the given month, according to Project Gutenberg}\label{tabpop}
\begin{tabular}{p{0.28\linewidth} | p{0.16\linewidth} | p{0.07\linewidth} | p{0.12\linewidth} | p{0.12\linewidth}}
\hline
\textbf{Book (Fiction genre)} & \textbf{Author} (Ground truth) & \textbf{Chunk}  & \textbf{Download} & 
\textbf{Wikipedia frequency}\\
\hline
Pride and Prejudice (Regency romance) & Jane Austen & 306 & 77,172 & 1,588 \\ 
\hline
Moby Dick (Adventure) & Herman Melville & 530 & 69,342 & 2,400 \\ 
\hline

Middlemarch (Historical) & George Eliot & 790 &  50,920 & 362 \\
\hline

The Adventures of Ferdinand Count Fathom (Gothic) & Tobias Smollett & 397 & 39,848 & 6 \\ 
\hline

The Expedition of Humphry Clinker (Epistolary) & Tobias Smollett & 371 & 38,788 & 46 \\ 
\hline

The Adventures of Roderick Random (Picaresque) & Tobias Smollett & 477 &  38,561 & 30 \\ 
\hline

History of Tom Jones (Picaresque) & Henry Fielding  & 864 & 37,986 & 128 \\ 
\hline

A Doll's House (Realist drama) & Henrik Ibsen & 67 & 29,637 & 939 \\ 
\hline

Crime and Punishment (Crime) & Fyodor Dostoevsky & 507 & 23,269 & 2,214 \\
\hline

Great Expectations (Gothic) & Charles Dickens & 922 & 19,251 & 2,489 \\ 
 \hline
\textbf{Total} &  & \textbf{5,231} & \textbf{424,774} & \\
 \hline
 \hline
\end{tabular}
\end{table}

\subsection{Zero-shot predictions by the \acrshort{llm}s}

Similarly to the annotation (or prediction) guideline for several case studies by \citet{ide2017introduction}, we designed the
lifecycle for the predictions as given in Figure \ref{annot_method}.
It begins with creating the chunks from the books.
We then prompt the \acrshort{llm}s for author attribution in a 3-fold loop, depending on if the output is empty, which occurred only with \acrshort{llama}-2.
After each iteration, the prompt is redesigned before it is fed to the \acrshort{llm} according to the points below (where \textit{{txt}} is the chunk of text).
The 2 follow-up prompts are designed with instruction because of the potential to improve performance, as shown in the literature \cite{wei2022chain,kojima2022large,10.1007/978-3-031-70442-0_5}.

\begin{enumerate}
    \item Who is the author of this text: \textit{`{txt}'}?
    \item \#\#\# Instruction: Following is a Question Answering task. As a helpful system, give a suitable response: Who is the author of this text: \textit{`{txt}'}?
    \item \#\#\# Instruction: Following is a Question Answering task. As a helpful system, give a suitable response: Who wrote this text: \textit{`{txt}'}?    
\end{enumerate}

After prediction, 162 chunks per book\footnote{We use all 67 total chunks for Ibsen} (as sample size) are randomly selected from each \acrshort{llm}-annotated set of chunks for \textbf{human evaluation} and post-processing.
The 162 (or 1,525 combined) chunks is based on the error margin of 7\%, a confidence interval of 95\% and because human evaluation of the total (5,231) is not very practical due to time constraint.
The post-processing refers to condensing the descriptive outputs into one-word labels.
These are: 1) the last name of the `\textit{correct}' author, 2) `\textit{incorrect}' or `\textit{others},' when it is an incorrect attribution, or 3) `\textit{unknown}', when the \acrshort{llm} does not know or there is still no output after the 3 prompts.
Effectively, these are the 3 high-level labels (correct, incorrect or unknown).





\section{Results and Discussion}
\label{res}

\begin{figure}[h!]
\centering
\includegraphics[width=0.5\textwidth]{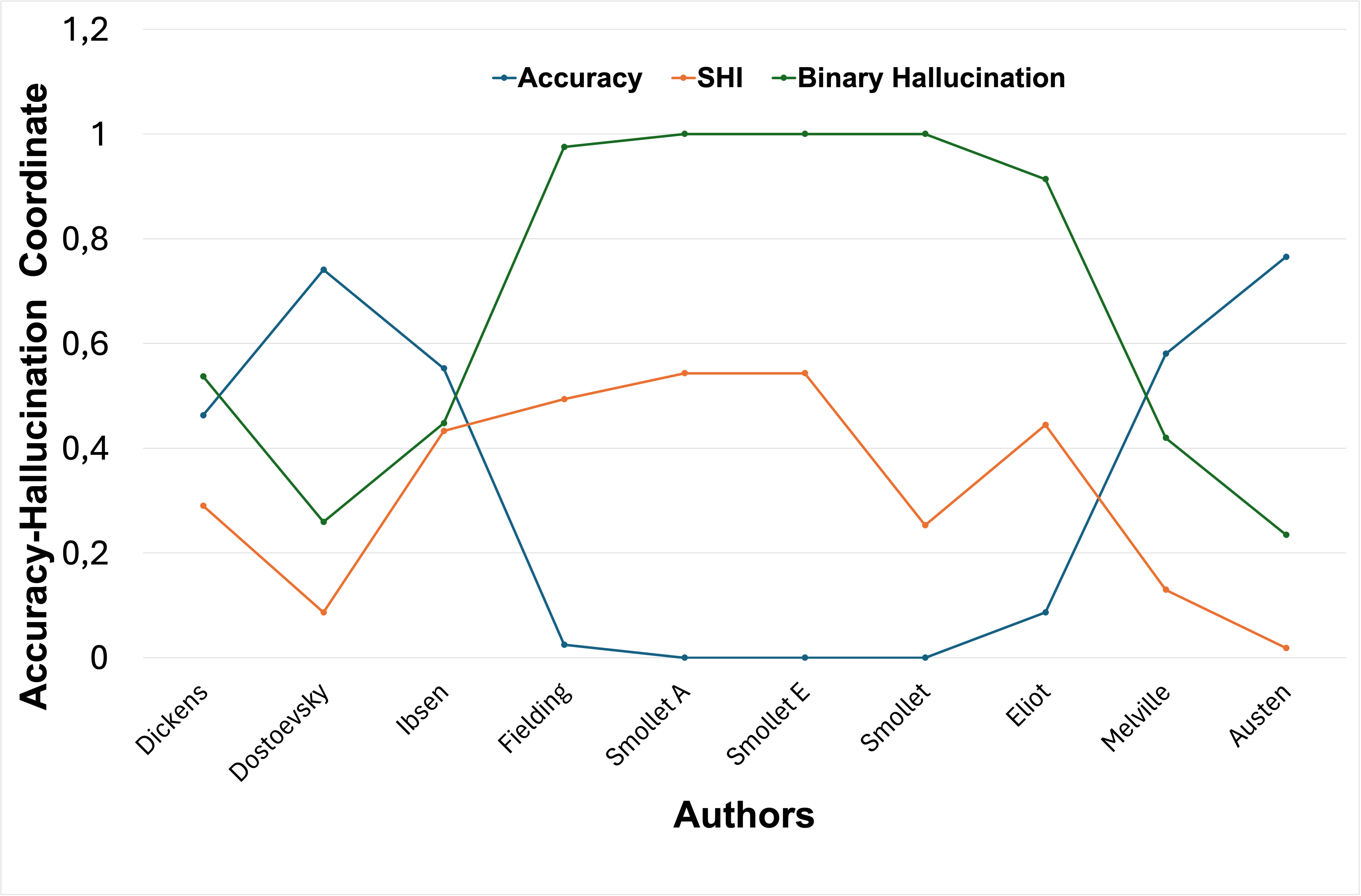}
\caption{Correlation of accuracy and hallucinations for Gemma-7B. (Smollett E- Expedition.. \& Smollett A- Adventures of Roderick..)} \label{fig3}
\end{figure}

\begin{figure}[h!]
\centering
\includegraphics[width=0.5\textwidth]{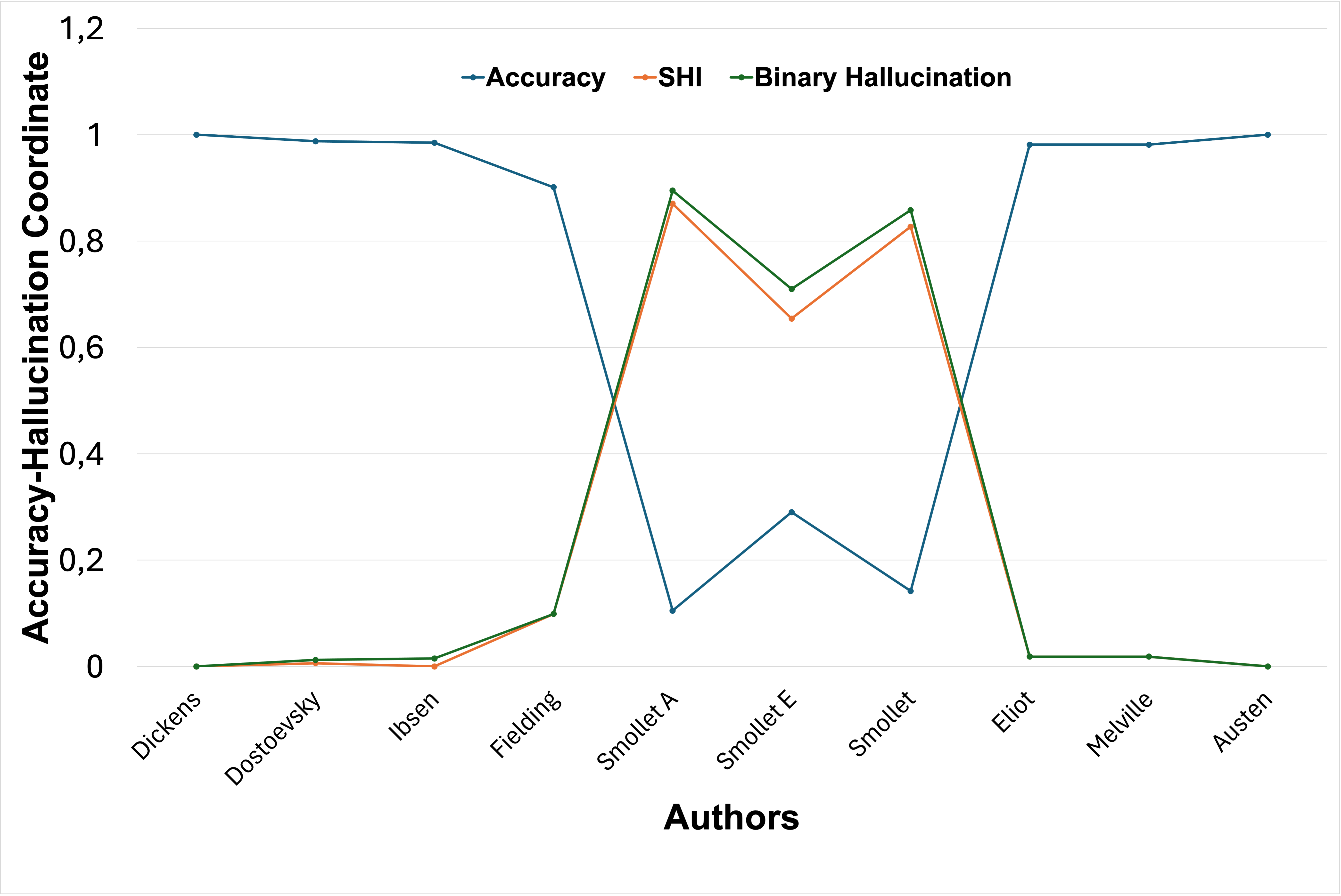}
\caption{Correlation of accuracy and hallucinations for Mixtral 8x7B. (Smollett E- Expedition.. \& Smollett A- Adventures of Roderick..)} \label{fig4}
\end{figure}

\begin{figure}[h!]
\centering
\includegraphics[width=0.5\textwidth]{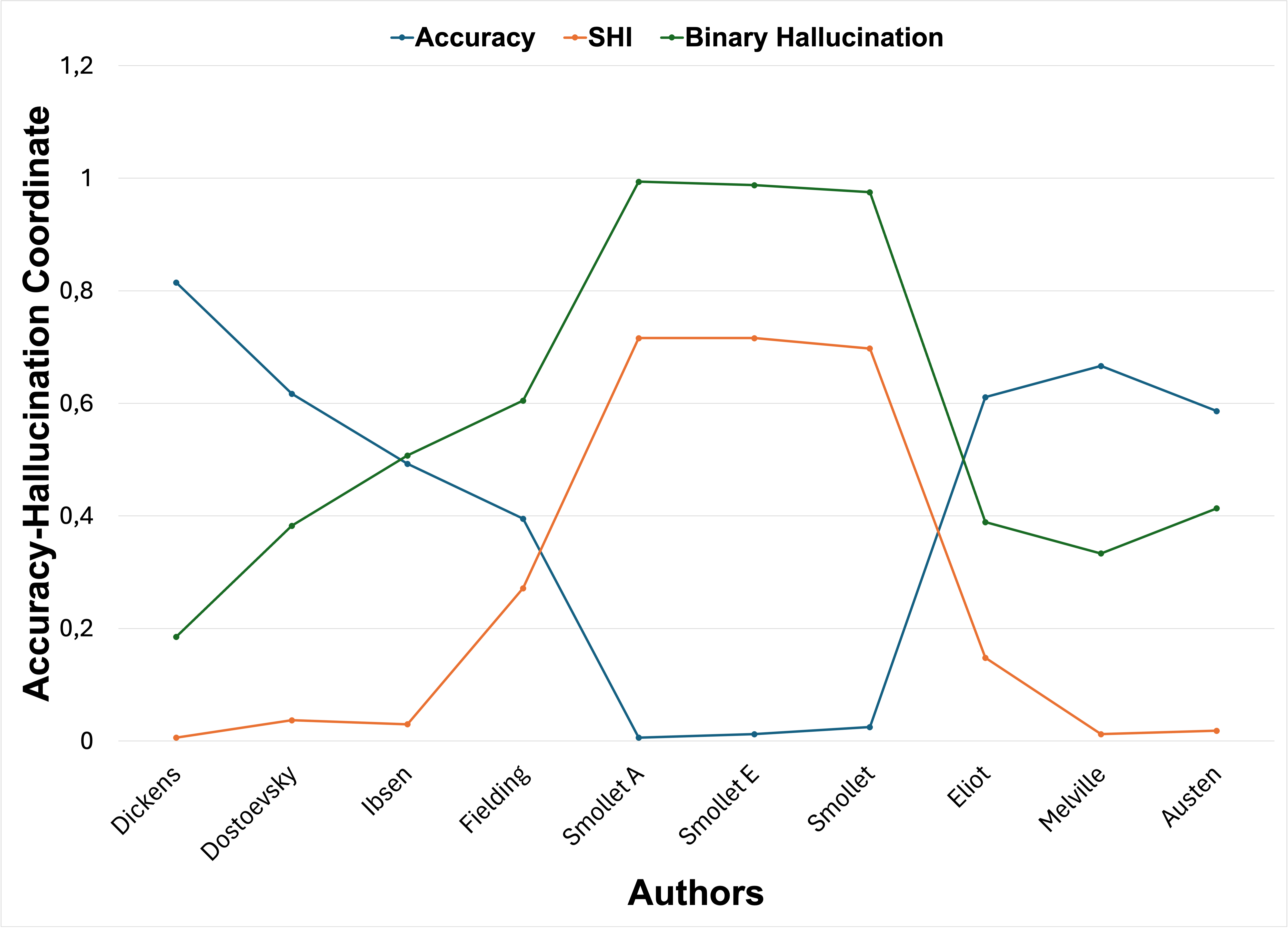}
\caption{Correlation of accuracy and hallucinations for \acrshort{llama}-2-13B (Smollett E- Expedition.. \& Smollett A- Adventures of Roderick..).} \label{fig5}
\end{figure}

Table \ref{resultstab} presents the average results for the 3 models over the 10 books.
Table \ref{tresults}\footnote{Binary hallucination is provided in the Figures due to space limitations} provides the details of the results of Table \ref{resultstab}.
Mixtral 8x7B has the best average performance across all scores, resulting in the best average accuracy and the lowest average \acrshort{shi}.
Gemma-7B has the lowest average accuracy and the highest average \acrshort{shi}.
The performance among the \acrshort{llm}s seem to follow the trend of their parameter sizes and this is no surprise.
The Pearson's correlation (\textit{r}) values in Table \ref{resultstab} are statistically significant, based on the p-value $<$ 0.00001 for the alpha of 0.05 for all the models.
We observe, based on \acrshort{shi}, that it is better for a model to admit it does not know an answer than to make a false attribution.
We also observe a strong negative correlation (\textit{r}) between accuracy and \acrshort{shi}, which is indicative of the fidelity of \acrshort{shi} in effectively scoring hallucinations.
Despite having the best average performance, Mixtral 8x7B hallucinates strongly on all the 3 books by \textit{Smollett}. 
This issue is observed for all the \acrshort{llm}s, as shown in
Figures \ref{fig3}, \ref{fig4}, and \ref{fig5} - Figures of correlation of accuracies and hallucinations.
Indeed, Smollett outliers are responsible for the large standard deviations in Table \ref{resultstab}.

\begin{table}[h!]
\small
\centering
\caption{Average results (and standard deviations) over the 10 books. Mixtral 8x7B performs best (c- correct, i- incorrect \& u- unknown). \footnotesize Smollett outliers responsible for large deviations.}\label{resultstab}
\begin{tabular}{p{0.13\linewidth} | p{0.10\linewidth} | p{0.06\linewidth} | p{0.06\linewidth} | p{0.06\linewidth} | p{0.10\linewidth} | p{0.13\linewidth}}
\hline
\textbf{Model} & \textbf{Acc}  $\uparrow$ & \textbf{c} $\uparrow$ & \textbf{i} $\downarrow$ & \textbf{u} $\downarrow$ & \textbf{\acrshort{shi}} $\downarrow$ & \textbf{\textit{r}} $\downarrow$\\
\hline
Gemma-7B & 0.309 (0.311) & 47 & 48 & 57 & 0.316 (0.186) & -0.8000\\
Mixtral 8x7B & \textbf{0.724} (0.369) & \textbf{110} & \textbf{40} & \textbf{2} & \textbf{0.263} (0.355) & \textbf{-0.9996}\\
\acrshort{llama}-2-13B & 0.421 (0.286) & 64 & 42 & 46 & 0.276 (0.301) & -0.9650\\
\hline
\end{tabular}
\end{table}

\begin{table}[h!]
\tiny
\centering
\caption{Detailed results for the \acrshort{llm}s}\label{tresults}
\begin{tabular}{p{0.13\linewidth} | p{0.15\linewidth} | p{0.08\linewidth} | p{0.07\linewidth} | p{0.07\linewidth} | p{0.07\linewidth} | p{0.08\linewidth}}
\hline
\textbf{Ground Truth} & \textbf{Model} & \textbf{Acc} $\uparrow$ & \textbf{c} $\uparrow$& \textbf{i} $\downarrow$ & \textbf{u} $\downarrow$ & \textbf{\acrshort{shi}} $\downarrow$\\
\hline
\multirow{3}{*}{Austen} & \acrshort{llama}-2-13B & 0.586 & 95 & 3 & 64 & 0.019 \\
 & Mixtral 8x7B & 1 & 162 & 0 & 0 & 0 \\
  & Gemma-7B & 0.765 & 124 & 3 & 35 & 0.019\\
  & &  & & &  & \\
\hline
\multirow{3}{*}{Melville} &  \acrshort{llama}-2-13B & 0.667  & 108 & 2 & 52 & 0.012\\
 & Mixtral 8x7B & 0.981 & 159 & 3 & 0 & 0.019\\
  & Gemma-7B & 0.580 & 94 & 21 & 47 & 0.130 \\
  & &  & & & &  \\
\hline
\multirow{3}{*}{Eliot} &  \acrshort{llama}-2-13B & 0.611 & 99 & 24 & 39 & 0.148 \\
 & Mixtral 8x7B & 0.981 & 159 & 3 & 0 & 0.019 \\
  & Gemma-7B & 0.086 & 14 & 72 & 76 & 0.444\\
  & &  & & &  & \\
\hline
\multirow{3}{*}{Smollett} &  \acrshort{llama}-2-13B & 0.025 & 4 & 113 & 45 & 0.698\\
 & Mixtral 8x7B & 0.142 & 23 & 134 & 5 & 0.827\\
  & Gemma-7B & 0 & 0 & 41 & 121 & 0.253 \\
  & &  & & & &  \\
\hline
\multirow{3}{*}{Smollett} &  \acrshort{llama}-2-13B & 0.012 & 2 & 116 & 44 & 0.716 \\
 & Mixtral 8x7B & 0.290 & 47 & 106 & 9 & 0.654 \\
(Expedition) & Gemma-7B & 0 & 0 & 88 & 74 & 0.543 \\
  & &  & & &  & \\
\hline
\multirow{3}{*}{Smollett} & \acrshort{llama}-2-13B & 0.006 & 1 & 116 & 45 & 0.716 \\
 & Mixtral 8x7B & 0.105 & 17 & 141 & 4 & 0.870 \\
(Adventures of Roderick)  & Gemma-7B & 0 & 0 & 88 & 74 & 0.543 \\
  & &  & & &  & \\
\hline
\multirow{3}{*}{Fielding} & \acrshort{llama}-2-13B & 0.395 & 64 & 44 & 54 & 0.272\\
 & Mixtral 8x7B & 0.901 & 146 & 16 & 0 & 0.098 \\
  & Gemma-7B & 0.025 & 4 & 80 & 78 & 0.494 \\
  & &  & & & &  \\
\hline
\multirow{3}{*}{Ibsen} & \acrshort{llama}-2-13B & 0.493 & 33 & 2 & 32 & 0.030 \\
 & Mixtral 8x7B & 0.985 & 66 & 0 & 1 & 0\\
  & Gemma-7B & 0.552 & 37 & 29 & 1 & 0.433 \\
  & &  & & &  & \\
\hline
\multirow{3}{*}{Dostoevsky} &\acrshort{llama}-2-13B & 0.617 & 100 & 6 & 56 & 0.037\\
 & Mixtral 8x7B & 0.988 & 160 & 1 & 1 & 0.006 \\
  & Gemma-7B & 0.741 & 120 & 14 & 28 & 0.086\\
  & &  & & & &  \\
\hline
\multirow{3}{*}{Dickens} & \acrshort{llama}-2-13B & 0.815 & 132 & 1 & 29 & 0.006 \\
 & Mixtral 8x7B & 1 & 162 & 0 & 0 & 0\\
  & Gemma-7B & 0.463 & 75 & 47 & 40 & 0.290 \\
  & &  & & & &  \\
 \hline
\end{tabular}
\end{table}

\begin{table}[h!]
\small
\centering
\caption{Empty chunk outputs after each iteration for \acrshort{llama}-2-13B.}\label{tabllama}
\begin{tabular}{p{0.5\linewidth} | p{0.08\linewidth} | p{0.08\linewidth} | p{0.08\linewidth}}
\hline
\textbf{Book (Author)} & \textbf{1st} & \textbf{2nd}  & \textbf{3rd}\\
\hline
Pride and Prejudice (Jane Austen) & 229 & 78 & 54\\ 
Moby Dick  (Herman Melville) & 403 & 86 & 50 \\ 
Middlemarch (George Eliot)& 280  & 106 & 55\\ 
The Adventures of Ferdinand Count Fathom (Tobias Smollett) & 98 & 30 & 9 \\
The Expedition of Humphry Clinker (Tobias Smollett) & 91 & 35 & 21 \\ 
The Adventures of Roderick Random (Tobias Smollett) & 105 & 27 & 12 \\ 
History of Tom Jones (Henry Fielding) & 374 & 167 & 96\\ 
A Doll's House (Henrik Ibsen) & 43 & 27 & 19\\ 
Crime and Punishment (Fyodor Dostoevsky) & 204 & 110 & 53 \\ 
Great Expectations (Charles Dickens) & 298 & 123 & 63 \\ 
 \hline
\end{tabular}
\end{table}

\subsection{Analysis}

\begin{figure}[h!]
\centering
\includegraphics[width=0.5\textwidth]{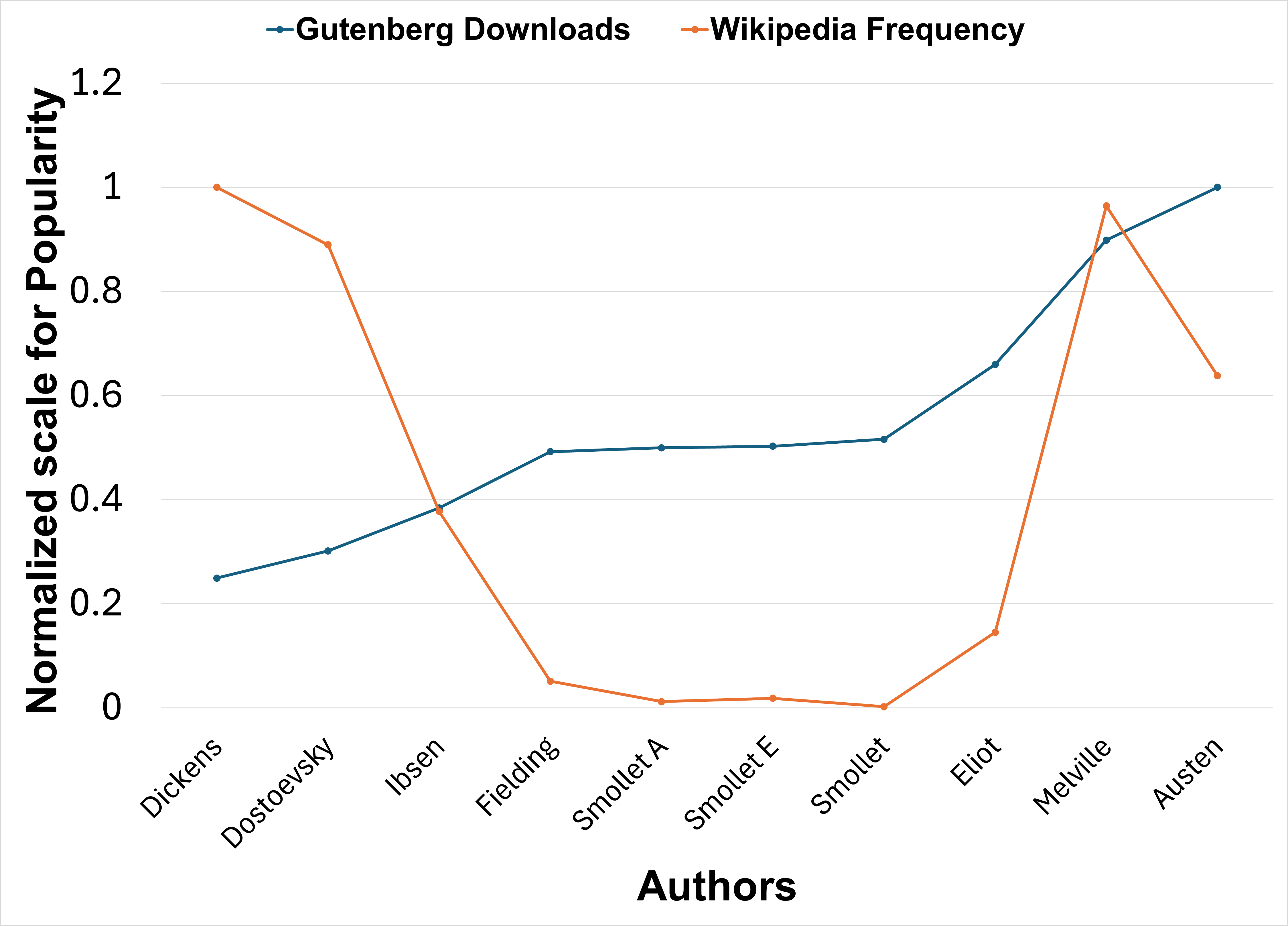}
\caption{Trendline of normalized downloads \& frequencies.} \label{fig2}
\end{figure}

Given that only \acrshort{llama}-2-13B used the 3-fold loop for predictions,
Table \ref{tabllama} shows the number of empty outputs at each iteration of its prompts.
From it, we observe that \textbf{adding instructions or prompt engineering helps} to obtain non-empty outputs.
Furthermore, to understand why hallucination is strongest for the books by Smollett and ascertain any general trend with the books (i.e. correlation between the accuracies and the popularity of the books), we plotted Figure \ref{fig2} after normalizing downloads and Wikipedia frequencies by their respective highest values.
We observe that the trendline of the \textbf{frequencies largely follows those of the accuracies of the 3 \acrshort{llm}s}, instead of the downloads, implying positive correlations with Pearson of 0.861, 0.68, and 0.82 for Gemma-7B, Mixtral 8x7B, and \acrshort{llama}-2-13B, respectively.
This confirms that the more the instances in the pretraining data, the better the performance on such a book.
Therefore, the 3 books by Smollett have the least frequencies and thus the worst results.
As expected, the number of downloads of a book has no meaningful relationship with performance.

\begin{figure*}[h!]
\centering
\includegraphics[width=1\textwidth]{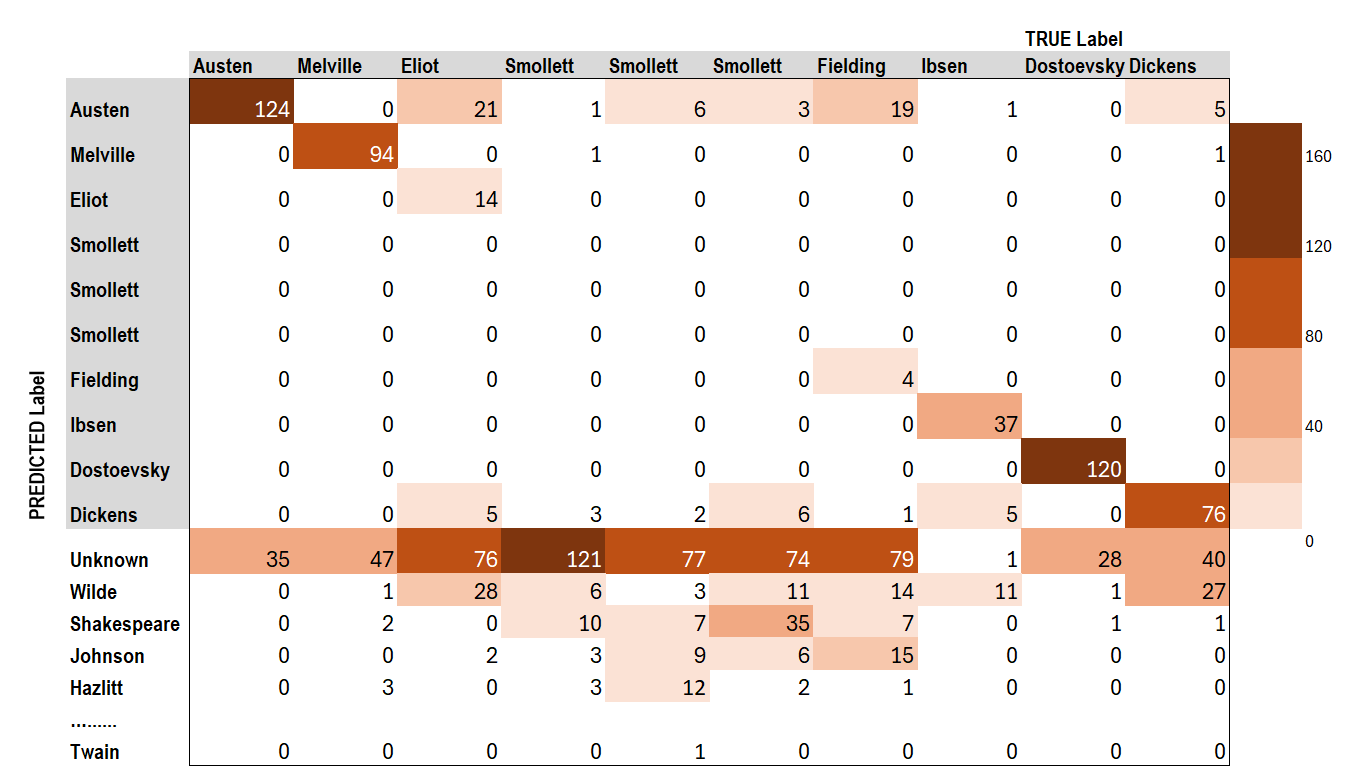}
\caption{Gemma-7B confusion matrix (Shows 'unknown', the 4 most false predictions and the least false prediction (out of 89)).} \label{fig6}
\end{figure*}

\begin{figure*}[h!]
\centering
\includegraphics[width=1\textwidth]{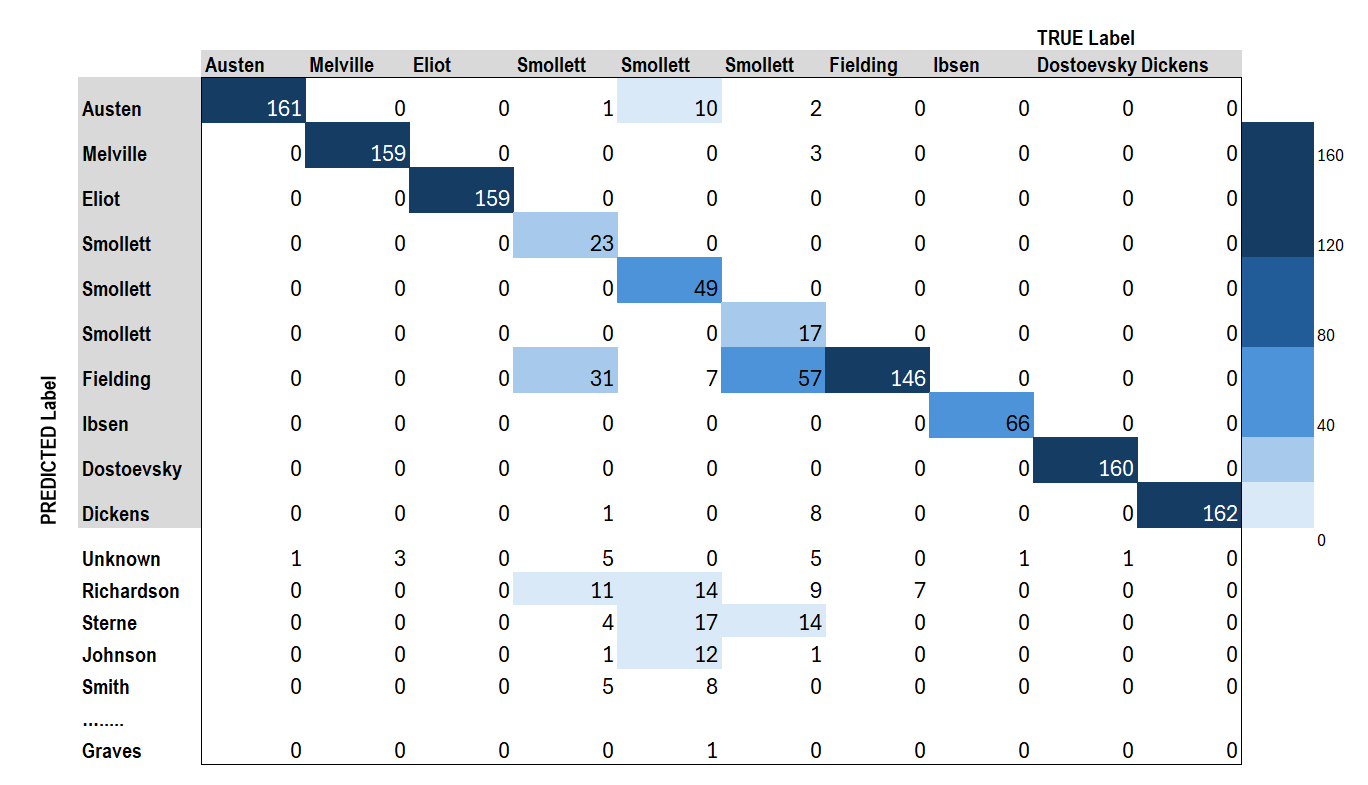}
\caption{Mixtral 8x7B confusion matrix (Shows 'unknown', the 4 most false predictions and the least false prediction (out of 100)).} \label{fig7}
\end{figure*}

\begin{figure*}[h!]
\centering
\includegraphics[width=1\textwidth]{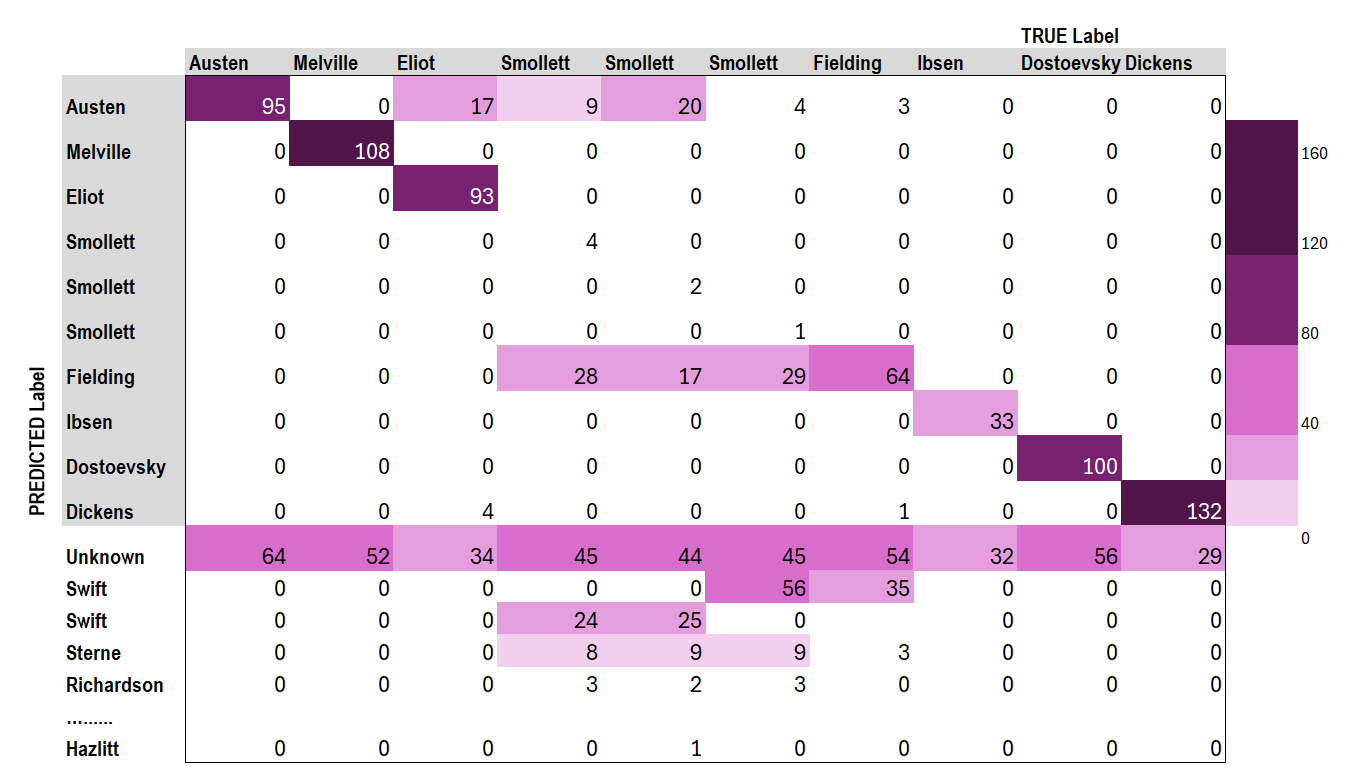}
\caption{\acrshort{llama}-2-13B confusion matrix (Shows 'unknown', the 4 most false predictions and the least false prediction (out of 80)).} \label{fig8}
\end{figure*}

Figures \ref{fig6}, \ref{fig7} and \ref{fig8} provide confusion matrices for Gemma-7B, Mixtral 8x7B and \acrshort{llama}-2-13B, resepctively.
They provide more detailed analyses of the predictions and errors made by the \acrshort{llm}s.
For Figure \ref{fig6}, Gemma-7B made a total of 89 different predictions of authors but has the worst pattern of correct predictions.
For Figure \ref{fig7}, Mixtral 8x7B made a total of 100 different predictions of authors but has the best pattern of correct predictions.
For Figure \ref{fig8}, \acrshort{llama}-2-13B made a total of 80 different predictions of authors and has a better pattern of correct predictions than Gemma-7B but worse pattern than Mixtral 8x7B.

\subsection*{Qualitative Examples}
Below we present two examples of responses to different chunks from \acrshort{llama}-2-13B.
The first example is a correct response and is post-processed (or labeled) as '\textit{dostoevsky}'.
The second is an incorrect response.

\begin{enumerate}
    \item Correct \\
    \textit{As a helpful system, I can provide information about the authorship of the text. The text is an excerpt from the novel "Crime and Punishment" by the Russian author Fyodor Dostoevsky. The novel was first published in 1866, and the excerpt provided is from the chapter titled "The Pro and the Contra." The text is written in the third person, using the narrative voice of the novel, and it describes a scene in which the character Svidrigaïlov is speaking to the character Avdotya Romanovna. Based on the language, style, and content of the text, it is clear that the author is Fyodor Dostoevsky.}
    
    \item Hallucination (Incorrect)\\
    \textit{I'm happy to help! The author of this text is Jonathan Swift, a renowned Irish satirist and writer. The text is an excerpt from his famous novel "Gulliver's Travels."}

    
\end{enumerate}




\section{Related Work}
\label{related}

\acrshort{llm}s have emerged as powerful \acrfull{gai} that can solve multiple tasks, including displaying emergent properties \cite{brown2020language,radford2019language}.
Some of these tasks are reasoning \cite{wei2022chain}, summarization \cite{cao2018faithful}, data-to-text \cite{wiseman-etal-2017-challenges}, and dialogue generation \cite{gehrmann-etal-2021-gem,nakamura-etal-2025-aurora}.
Author attribution for \acrshort{gai} is also emerging as an area of study \cite{10438784,adewumi2023procot}.
Apparently, one of the attendant problems with the power of \acrshort{llm}s is hallucination \cite{ji2023survey,pettersson2024generative}.

Following the introduction of stylometric methods to author attribution \cite{ mendenhall1901menchanical,malyutov2006authorship}, today different \acrfull{ml} methods are being applied to the different forms of the task \cite{argamon2008interpreting,koppel2009computational}.
\citet{10.1007/978-3-030-49161-1_22} used a character-level recurrent neural network (RNN) and a multi-headed classifier in closed-set attribution.
They also considered Universal Language Model Fine-Tuning (ULMFiT), Embeddings from Language Models (ELMo), Generative Pretrained Transformer 2 (GPT-2), and \acrfull{bert}.
\citet{wang2021integrating} used the \acrfull{roberta} model in the author attribution of short texts.
\citet{hicke2023t5} finetuned Falcon and Pythia \acrshort{llm}s to generate predictions of authors for texts of plays written in the 16th century, in addition to using logistic regression and Support Vector Machine.
These works did not consider the hallucination problem or try to estimate the size of the problem in a systematic way.

\section{Conclusion}
\label{conclude}

We showed, in this work, that recent \acrshort{llm}s are powerful but they still suffer from high hallucinations in some cases when it comes to author attribution.
Our newly introduced hallucination metric (\acrshort{shi}) demonstrates fidelity in providing an effective score for hallucination in a given task.
This new metric has a strong negative correlation with prediction accuracy.
We strongly believe that adequately gauging a problem will provide the opportunity to adequately tackle it.
As future work, it can be interesting to evaluate closed \acrshort{llm}s, such as ChatGPT and explore more books and datasets.

\section*{Acknowledgments}
This work is supported by the Wallenberg AI, Autonomous Systems and Software Program (WASP), funded by Knut and Alice Wallenberg Foundations and counterpart funding from Luleå University of Technology (LTU).
We thank the anonymous reviewers for their valuable feedback in improving this paper.

\bibliographystyle{acl_natbib}
\bibliography{ranlp2023}


\end{document}